\documentclass[final,3p,times]{elsarticle}

\usepackage{amssymb}
\usepackage{amsmath}
\usepackage{algorithm}
\usepackage{algorithmic}
\usepackage{subfigure}
\usepackage{graphicx}
\usepackage{float}
\usepackage{subfig}
\date{}
\usepackage{setspace}
\usepackage{threeparttable}
\usepackage{multirow}
\usepackage{booktabs} 
\usepackage[figuresright]{rotating}
\usepackage{adjustbox}
\usepackage[normalem]{ulem}

\begin{document}

\begin{frontmatter}
\title{Depth-Sensitive Soft Suppression with RGB-D Inter-Modal Stylization Flow  for Domain Generalization Semantic Segmentation}

\author[2,3]{Binbin Wei}
\cortext[cor1]{Corresponding author at: College of Electronics and Information Engineering, Shenzhen University, Shenzhen, 518060, China. e-mail: wzou@szu.edu.cn (Wenbin Zou). Binbin Wei and Yuhang Zhang contributed equally to this work. e-mail: 2200432094@email.szu.edu.cn; zhangyuhang2019@email.szu.edu.cn; stian@szu.edu.cn; seabearlmx@gmail.com; 2250432002@email.szu.edu.cn}
\author[2,3]{Yuhang Zhang}
\author[1,2,3]{Shishun Tian}
\author[4]{Muxin Liao}
\author[2,3]{Wei Li}
\author[1,2,3]{Wenbin Zou\corref{cor1}}

\affiliation[1]{organization={Guangdong Key Laboratory of Intelligent Information Processing},
	addressline={Shenzhen University}, 
	city={China},
	postcode={518060}, 
	state={Guangdong},
	country={shenzhen}}

\affiliation[2]{organization={Shenzhen Key Laboratory of Advanced Machine Learning and Applications},
	addressline={Shenzhen University}, 
	city={China},
	postcode={518060}, 
	state={Guangdong},
	country={shenzhen}}

\affiliation[3]{organization={College of Electronics and Information Engineering},
	addressline={Shenzhen University}, 
	city={China},
	postcode={518060}, 
	state={Guangdong},
	country={shenzhen}}

\affiliation[4]{organization={School of Computer and Information Engineering},
	addressline={Jiangxi Agricultural University}, 
	city={China},
	postcode={330045}, 
	state={Jiangxi},
	country={Nanchang}}

\begin{abstract}

Unsupervised Domain Adaptation (UDA) aims to align source and target domain distributions to close the domain gap, but still struggles with obtaining the target data. Fortunately, Domain Generalization (DG) excels without the need for any target data. Recent works expose that depth maps contribute to improved generalized performance in the UDA tasks, but they ignore the noise and holes in depth maps due to device and environmental factors, failing to sufficiently and effectively learn domain-invariant representation. Although high-sensitivity region suppression has shown promising results in learning domain-invariant features, existing methods cannot be directly applicable to depth maps due to their unique characteristics. Hence, we propose a novel framework, namely Depth-Sensitive Soft Suppression with RGB-D inter-modal stylization flow (DSSS), focusing on learning domain-invariant features from depth maps for the DG semantic segmentation. Specifically, we propose the RGB-D inter-modal stylization flow to generate stylized depth maps for sensitivity detection, cleverly utilizing RGB information as the stylization source. Then, a class-wise soft spatial sensitivity suppression is designed to identify and emphasize non-sensitive depth features that contain more domain-invariant information. Furthermore, an RGB-D soft alignment loss is proposed to ensure that the stylized depth maps only align part of the RGB features while still retaining the unique depth information. To our best knowledge, our DSSS framework is the first work to integrate RGB and Depth information in the multi-class DG semantic segmentation task. Extensive experiments over multiple backbone networks show that our framework achieves remarkable performance improvement.

\end{abstract}

%


\end{frontmatter}


\section{Introduction}
Semantic segmentation \cite{ye2023multi},\cite{zhou2024pggnet},\cite{zhu2025fanet},  making each pixel in an image categorized, has achieved remarkable success with the development of deep neural networks \cite{chen2018encoder}, \cite{xie2021segformer}. However, the performance of models degrades significantly when there is a distribution discrepancy between the training and testing data. To mitigate this issue, Unsupervised Domain Adaptation (UDA) and Domain Generalization (DG) have been proposed to improve the cross-domain generalization capabilities of models. UDA\cite{liao2022exploring},\cite{shu2024adatriplet},\cite{zhang2025global} aims to minimize the domain gap by leveraging annotated source data and unlabeled target data during training.  However, prior access to target domains is often impractical, and performance degradation still occurs in unseen domains. Domain Generalization (DG) is well-suited to address these issues. In the setting of the DG task, the model aims to extract domain-invariant information by using only source data on the training stage, and is capable of generalizing well to any unseen target domains\cite{liao2024class}, \cite{zhang2023fine}.

\par Among the existing methods, depth information assistance is crucial for learning domain-invariant representation effectively\cite{wang2021domain},\cite{hu2023multi}, \cite{yang2024micdrop}. In this paper, we focus on utilizing depth information to achieve depth-invariant representation learning. By reviewing previous research, two significant insights about the depth map are concluded as follows.

\par
\begin{figure}[h]
	\centering
	\includegraphics[width=1\textwidth]{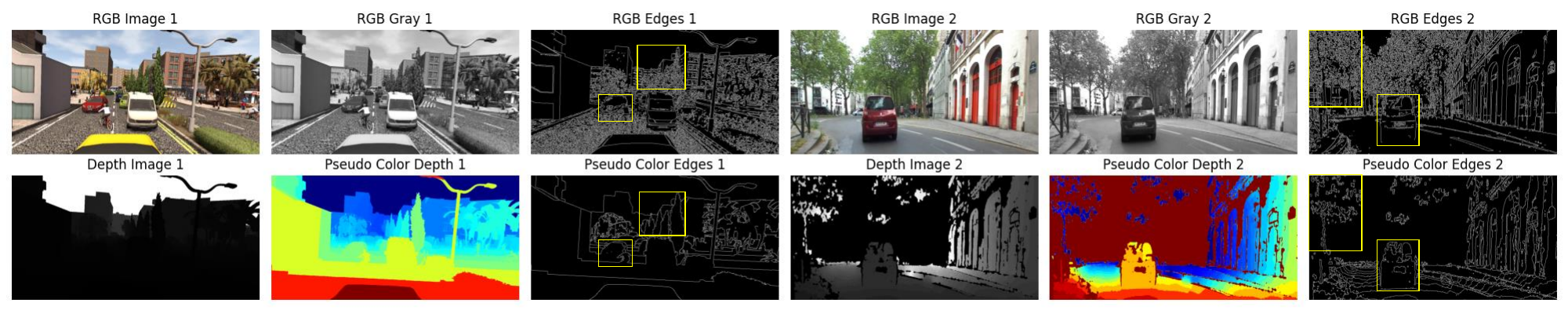}
	\caption{Comparative visualization of RGB and depth images and their edge detection results for the synthetic SYNTHIA dataset\cite{ros2016synthia} (image pair 1) and the real-world InfraParis dataset\cite{franchi2024infraparis} (image pair 2). The depth images are colorized for easier visual comparison, while the original depth images are grayscale single-channel images.}
	\label{depth_advantage}
\end{figure}

 First, RGB-D-based domain generalization semantic segmentation (DGSS) method under multi-class complex environment has not been fully explored yet. The effectiveness of depth map reflecting the shape and spatial position of objects, has been verified in some UDA methods. Some works \cite{lee2018spigan}, \cite{chen2019learning}, \cite{vu2019dada} infer the depth maps from color images with the usage of depth estimation tasks, while some works \cite{wang2021domain},\cite{hu2023multi}, \cite{yang2024micdrop}  innovatively feed the depth maps and RGB data into the model simultaneously. They all take advantage of the conclusion that depth map is one type of domain-invariant feature. To better understand this conclusion, related visualization is shown in Fig. \ref{depth_advantage}. As shown in the first three columns of Fig. \ref{depth_advantage}, edge comparison shows that the depth map presents the shape and spatial position of the object more clearly with less texture interference, highlighting that the depth map includes rich domain-invariant information. Although the depth map has been explored in UDA field, it still has a wide room for
investigation in the DG task. To our best knowledge, our previous work
DGFD\cite{liu2024segment} is the only work focusing on the RGB-D DGSS task. 
However, it is limited to two categories (“Road” and “Background”) and cannot handle more realistic complex traffic situations. In addition, DGFD relies on the fusion of surface normal maps requiring additional and time-consuming processing from the depth map. Therefore, research on RGB-D DGSS task under multi-class complex scenes is necessary and urgent.

\par Second, depth maps are not always perfect, especially in the real world. Different from the above RGB-D methods in cross-domain learning tasks that coarsely consider the depth information as the reliable domain-invariant information, we hold that the depth map cannot fully represent domain-invariant information as it still has some noises like holes and errors limited by equipment accuracy and environment affect. As illustrated in the last three columns of Fig. \ref{depth_advantage}, the real-world depth map shows fragment and discontinuous, which presents different from synthetic depth map with smooth property as shown in the first three columns of Fig. \ref{depth_advantage}. Therefore, how to alleviate these noise and learn reliable domain-invariant information from the depth map should be focused. 

Recent studies\cite{choi2021robustnet}, \cite{xu2022dirl} have shown that the RGB-based high-sensitivity region suppression allows the model to focus on weakly sensitive regions containing numerous domain-invariant features, leading to improved generalization performance. This technique is composed of two main components: sensitivity detection and high-sensitivity suppression. However, high-sensitivity suppression for depth maps has not been investigated yet, and existing RGB-based methods are not suitable for depth maps due to the following reasons:

\textbf{\textit{(1) In the sensitivity detection step, applying common augmentation on depth images could lead to distortions or the loss of critical information.}} The existing methods\cite{choi2021robustnet}, \cite{xu2022dirl} often rely on widely used data augmentation strategies (like color transformations or Gaussian blurs) to perturb RGB data and measure feature sensitivity based on the differences between the original and perturbed features. It is important to note that RGB images are characterized by richer color and texture information, whereas depth images contain the geometric structure of the scene. Given that the geometric information in depth images primarily captures object distances and spatial relations, simple blurring and photometric transformations are predominantly aimed at altering color and texture. Therefore, the use of common augmentation strategies in depth maps could lead to distortions or
the loss of critical information.

\textbf{\textit{(2) The existing channel-based high-sensitivity suppression, which can also be viewed as a form of coarse sensitivity suppression, fails to capture class-specific and spatial structural sensitivity.}} When it comes to high-sensitivity suppression, sensitivity suppression usually focuses on channels \cite{xu2022dirl}, as each channel in RGB images contains distinct color information, allowing for effective processing at the channel level. However, depth images have only one channel, and this channel primarily contains distance information, lacking the richness of color features.
Instead, the spatial structure, which encompasses geometric shapes and the relative positions of objects, plays a more critical role. Aggregating information from surrounding pixels can help mitigate the impact of noise. 
Moreover, the feature distributions of each class are often different, and the sensitivity to differences varies across classes. Common sensitivity suppression strategies are typically global, which may result in the neglect of sensitive features in some classes.

\textbf{\textit{(3) The existing hard suppression struggles to capture continuous variations and detailed information of sensitive regions due to its binarization operation.}} High-sensitivity hard suppression distinguishes sensitive and non-sensitive regions using a sensitivity threshold, and then binarizes the sensitive areas as 0 or 1. This extreme operation ignores the gradual variations in sensitivity across spatial regions, leading to information loss. Especially at the boundary between the highly sensitive area and the non-sensitive area, it is easy to introduce artifacts or distortion, and it is impossible to accurately capture the detailed information of the transition area. Moreover, this strategy lacks flexibility, making it challenging to adapt to diverse data distributions.

\par Based on the above analysis, a high-sensitivity soft suppression method tailored for depth maps is proposed for DGSS. Specifically, first, to increase the diversity of depth map data for sensitivity detection, we propose the RGB-D inter-modal stylization flow to stylize depth features without auxiliary datasets, providing efficiency and greater control. Second, to solve the problem that existing methods cannot accurately capture class-specific sensitivity and spatial sensitivity features, we propose the class-wise soft spatial sensitivity suppression. This focuses on maximizing the extraction of non-sensitive features for each class, thereby learning more domain-invariant features to improve model performance. Finally, we propose the RGB-D soft alignment loss to ensure that the stylized depth map retains information about partial RGB features, while preserving the inherent properties of the original depth map. 
\par In conclusion, the contributions of our work are summarized as follows:
\begin{itemize}
	\item{Based on the insight that depth maps cannot always provide reliable domain-invariant features, we propose a novel framework for the DGSS task, namely Depth-Sensitive Soft Suppression with RGB-D inter-modal stylization flow. To the best of our knowledge, this is the first study to explore RGB-D-based DGSS in multi-class complex environments.}
	\item{To improve data diversity for sensitivity detection, we propose an RGB-D inter-modal stylization flow and an RGB-D soft alignment loss. The former cleverly uses RGB as the stylized source to generate stylized depth maps, while the latter allows the stylized depth maps to retain partial RGB information while retaining the unique characteristics of depth maps.}
	\item{To enhance non-sensitive features that contain more domain-invariant information, we propose a class-wise soft spatial sensitivity suppression, overcoming the issue that previous methods cannot capture category-specific and spatial structure sensitivity.}
	\item{The proposed framework demonstrates superior performance compared to state-of-the-art methods on multiple benchmarks, achieving notable success in distinguishing roads and sidewalks, segmenting small objects, and also in low-light conditions.}
\end{itemize}

\section{Related Work}
\subsection{RGB-D Semantic Segmentation}
Depth maps, reflecting the physical distance from the optical center of camera to target objects, can provide additional geometric clues to complement the information from RGB images. Pixel-wise summation and direct concatenation into a four-channel image are common and simple methods, such as FuseNet\cite{hazirbas2017fusenet}, LDFNet\cite{hung2019incorporating}, but the performance often is limited. More effectively fusion manners are proposed including the attention mechanisms, signal filtering, gating units et al. ACNet\cite{hu2019acnet} utilizes the channel attention module to separately reweight features extracted from two modality features before summing, while CANet\cite{zhou2022canet} and CMPFFNet\cite{zhou2023cmpffnet} turn to exploit the joint spatial attention to get more robust features. SA-Gate\cite{chen2020bi} filters unnecessary information and corrects clues by splitting and gathering features, while GC-FFM\cite{zhao2023cross} uses gated cross-attention to merge two modality features. At the same time, transformers driven methods \cite{liu2021swinnet}, \cite{zhang2023cmx} are gradually used in research and have achieved surprise results. CMX\cite{zhang2023cmx} firstly designs a unified fusion transformer model named RGB-X, which can be leverage depth, thermal, polarization, event and LiDAR as additional modality. Quantitative experiments show surprising performance in all five modes. However, most of them are limited to research on indoor semantic segmentation methods based on supervised learning.

\subsection{UDA for semantic segmentation}
UDA transfers knowledge from labeled source domain images to target domain images that lack labels\cite{zhang2018unsupervised},\cite{zhang2023hybrid}, with adversarial training and self-training commonly employed to achieve domain-invariant knowledge. Adversarial training narrows the distribution gap between the two domains at various levels. At the image level, techniques like CyCADA\cite{hoffman2018cycada}, BDL\cite{li2019bidirectional}, and FDA\cite{yang2020fda} transform the appearance of images to make them look more similar across domains. At the feature level, methods such as DAST\cite{yu2021dast} leverage the attention mechanism to assign higher weights to the hard-to-adapt regions. At the output level, models like AdaptSegNet \cite{tsai2018learning}, FADA \cite{wang2020classes}, HDL \cite{zhang2023hybrid}, and ASANet\cite{zhou2020affinity} focus on ensuring that predictions remain consistent across both domains. On the other hand, self-training methods \cite{pan2020unsupervised}, \cite{mei2020instance}, \cite{zheng2021rectifying} boost the model by generating pseudo-labels for target domain in a self-supervised manner. However, UDA methods still have performance degradation in unseen domains.

\subsection{UDA for RGB-D semantic segmentation}
In RGB-D-based unsupervised domain adaptation (RGB-D UDA), depth information is typically used either as an auxiliary task or as an additional input modality. When used as an auxiliary task, depth helps enhance the performance of the main task by offering geometric knowledge like SPIGAN \cite{lee2018spigan}, DADA \cite{vu2019dada}. When depth is treated as an additional modality, both RGB and depth data are used together during training, enabling the model to capture richer features from multiple dimensions. CorDA \cite{wang2021domain} utilizes depth information from both domains, with the target depth is generated off-line. In contrast, MMADT \cite{hu2023multi} directly feeds depth and RGB into model as input during training and testing using a multi-modal strategy. To fully learn the joint features, MICDrop \cite{yang2024micdrop} designs a mask structure that allows RGB and Depth to capture features from only one modality at the same location during training. Considering that producing depth information in synthetic environments is nearly cost-free, whereas collecting depth maps in real-world scenarios is significantly more challenging. MDKT \cite{liu2024transferring} innovatively applies adversarial learning to transfer multimodal knowledge into a single modality. 

\subsection{DG for semantic segmentation}
Domain generalization methods, which optimize the model only with the labeled source data, can be maintain boost result on unseen domains overcoming the drawback of UDA methods. Domain randomization (DR) and feature normalization (FN) are two widely adopted techniques in existing DG methods. The primary reason of generating domain gap is the style gap between source and target domains. Some approaches stylize the texture and color of source images by image translation or adaptive instance normalization (AdaIN) such as DRPC \cite{yue2019domain}, GLTR \cite{peng2021global}, WildNet \cite{lee2022wildnet}. Several methods have pioneered a shift towards stylizing images in the frequency domain. FSDR \cite{huang2021fsdr} only randomizes the domain variant features after decoupling. Pasta \cite{chattopadhyay2023pasta} proposes a proportional amplitude spectrum training augmentation strategy to randomize images in the Fourier domain, where higher frequency components in the amplitude spectrum are more disturbed than lower frequency components.
FN methods learn the domain-agnostic knowledge by removing the domain-specific information such as IBN \cite{pan2018two}, ISW \cite{choi2021robustnet}. Some methods introduce the idea about feature sensitivity, which pay attention to the feature itself, such as SiamDoGe \cite{wu2022siamdoge}, DIRL \cite{xu2022dirl}.

\section{Approach}
\subsection{Preliminaries}
Sensitivity suppression is often used in cross-domain learning to help the model focus on weakly sensitive areas that contain numerous domain-invariant features. It is usually divided into two steps. Initially, sensitivity detection is carried out. Let $X\in \mathbb{R}^{C, H, W} $ denotes a feature map, where $C$, $H$, and $W$ represent the number of channels, and the height and width of the feature map, respectively. Often, this technique creates an additional augmented view by perturbing or transforming the original data, followed by a comparison of the two. The difference map obtained indicates sensitive information, where a greater difference signifies that the area is more susceptible to perturbations. The detection process is formalized as follows.
\begin{equation}
	\label{deqn_ex1a}
	\hat{X} = f(X)
\end{equation}
\begin{equation}
	\label{deqn_ex1a}
	\bigtriangleup = D(X, \hat{X})
\end{equation}
where $f$, $\hat{X}$, $D(\cdot)$, and $\bigtriangleup$ denote the augmentation operations, the augmented feature map, the method for calculating the difference and the difference between the two, respectively. Afterward, pixels are evaluated against the defined threshold to identify which are sensitive or non-sensitive. Generally, values above the threshold are designated as sensitive, while those below it are labeled non-sensitive. As shown below, sensitive pixels are set to 1 and insensitive pixels are set to 0, so that a sensitivity matrix $M$ can be obtained.
\begin{equation}
	\label{deqn_ex1a}
	\text{M} = 
	\begin{cases} 
		1, & \text{if } \Delta_{p} > \alpha \\ 
		0, & \text{otherwise} 
	\end{cases}
\end{equation}
where $\alpha $ denotes the established threshold, while $p$ represents a pixel on the difference feature map. Then the sensitive alignment loss $\mathcal{L} $ is used to close the distinctiveness between $\bigtriangleup$ and $M$ by performing the constraint function $F(\cdot)$, which can be generalized as:
\begin{equation}
	\label{deqn_ex1a}
	\mathcal L = F(M, \bigtriangleup)
\end{equation}
However, this technique, high-sensitivity hard suppression, is overly reliant on threshold selection, making it quite subjective. It binarizes the difference matrix (1 for values greater than the threshold, 0 for values less), which results in the loss of fine-grained information. High-sensitivity hard suppression is particularly coarse in scenarios with strong sensitivity and continuous values, making it difficult to capture the true feature distribution. Additionally, it may inadvertently suppress features that positively contribute to model performance.
\begin{figure*}[t]
	\centering
	\includegraphics[width=1\textwidth]{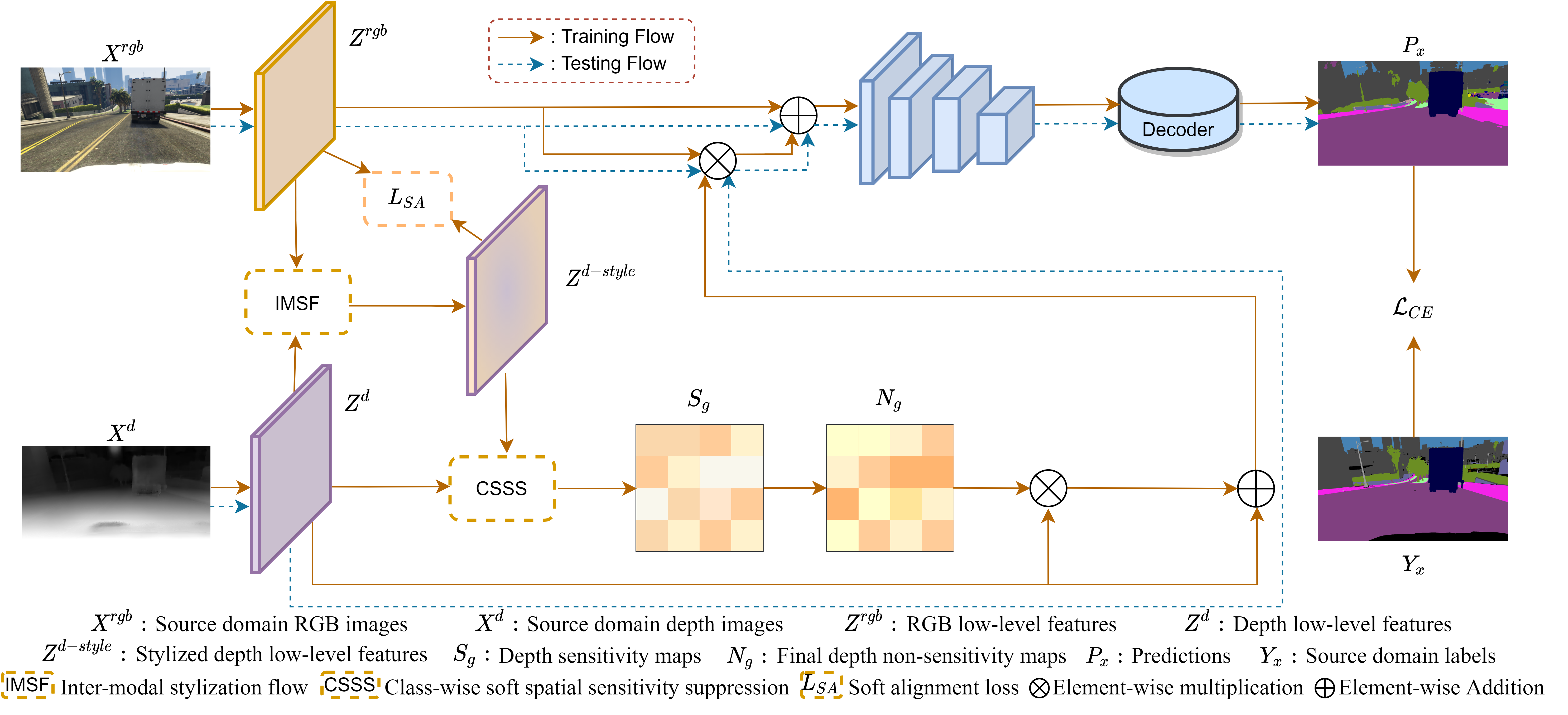}
	\caption{Flowchart of the proposed framework including the RGB-D inter-modal stylization flow, the class-wise soft spatial sensitivity suppression, and the RGB-D soft alignment loss. Only in the training phase are the proposed components applied to facilitate the generation of domain-invariant representations.}
	\label{overview}
\end{figure*}
\subsection{Overview}
As illustrated in Fig. \ref{overview}, the proposed framework is composed of three designed parts: the RGB-D inter-modal stylization flow, the class-wise soft spatial sensitivity suppression, and the RGB-D soft alignment loss. Formally, $D_{s} ={(X^{rgb}, X^{d}, Y_{x})}_{i=1}^{N} $ is used in training process, where $i \in N$, $X^{rgb}$ denotes an RGB image, $X^{d}$ denotes a depth image, and $Y_{x}$ denotes the label corresponding to the RGB image. The proposed framework enhances the generalization ability in multi-modal fusion manner, and the ultimate objective function of our approach is described as: 
\begin{equation}
	\label{deqn_ex1a}
	\mathcal{L} = \mathcal{L}_{CE} + \mathcal{L}_{SA}
\end{equation}
where \( \mathcal{L}_{CE}\) and $\mathcal{L}_{SA} $ represent the cross entropy loss, commonly adopted by other mainstream methods\cite{liao2024class}, \cite{choi2021robustnet}, and the proposed RGB-D soft alignment loss, respectively.

\subsection{RGB-D inter-modal stylization flow}
As mentioned in the Preliminaries, sensitivity suppression includes detection and suppression. A major module of sensitivity detection is the perturbation module, which generates perturbation features. Then the sensitivity is obtained by calculating the difference between the original features and the perturbation features. Directly using common augmentation strategies on the depth map to generate perturbed views may lead to the distortions or loss of critical information in the depth map. At the same time, considering the noise in the depth map, RGB image can provide or supplement the information that the depth map cannot capture or lose, such as detailed textures, edges of objects, etc. Therefore, it is a very intuitive approach to supplement the depth map with RGB information during the depth map stylization process. Inspired by domain flow \cite{gong2019dlow} and SiamDoGe \cite{wu2022siamdoge}, the RGB-D inter-modal stylization flow, which denotes the intermediate modal for domain generalization, is proposed to find shared representations for RGB and depth images within the same feature space, as shown in Fig. \ref{IMF}. By adjusting the mean and variance of the feature maps, the feature distribution discrepancies between different modalities can be reduced. In fact, it is to combine the semantic information of RGB and the spatial information of the depth map to stylize the depth map and generate diverse training data, which helps the model learn domain-invariant features. To begin with, the RGB-D inter-modal stylization flow is defined using their feature statistics as follows:
\begin{figure}[t]
	\centering
	\includegraphics[width=0.8\textwidth]{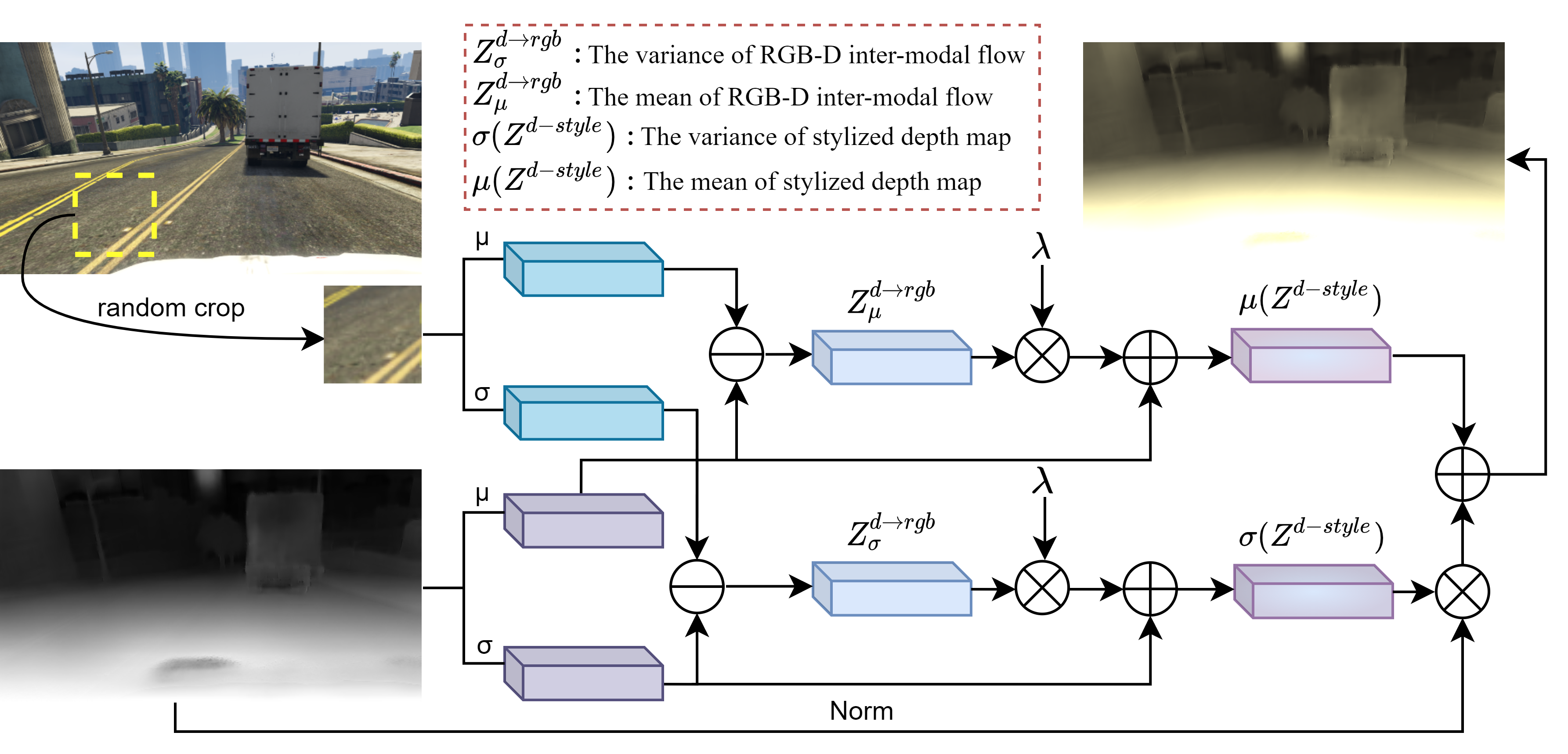}
	\caption{The illustration of the proposed RGB-D inter-modal stylization flow. The size of random cropping is typically set to 64$\times$64, and the random value $\lambda$ can only be between 0 and 1.}
	\label{IMF}
\end{figure}

\begin{equation}
	\label{deqn_ex1a}
	Z^{d\rightarrow rgb}_{\mu} = \mu(C(Z^{rgb})) - \mu(Z^{d})
\end{equation}
\begin{equation}
	\label{deqn_ex1a}
	Z^{d\rightarrow rgb}_{\sigma} = \sigma(C(Z^{rgb})) - \sigma(Z^{d})
\end{equation}
where $C$ denotes the random crop operation with the size of 64$\times$64, used to increase the diversity of flow as much as possible. The calculated inter-modal stylization flow is subsequently employed to randomize $Z^{d}$ in the following manner:
\begin{equation}
	\label{deqn_ex1a}
	\mu(Z^{d-style}) = \mu(Z^{d}) + \lambda Z^{d\rightarrow rgb}_{\mu}
\end{equation}
\begin{equation}
	\label{deqn_ex1a}
	\sigma(Z^{d-style}) = \sigma(Z^{d}) + \lambda Z^{d\rightarrow rgb}_{\sigma}
\end{equation}
\begin{equation}
	\label{deqn_ex1a}
	Z^{d-style} = \sigma(Z^{d-style})Z^{d-nor}+ \mu(Z^{d-style})
\end{equation}
where a random value $\lambda \in [0,1]$ is assigned to control the degree of style randomization, $Z^{d-nor}$ and $Z^{d-style}$ are normalized and stylized depth features, respectively.

\begin{figure*}[t]
	\centering
	\includegraphics[width=1.0\textwidth]{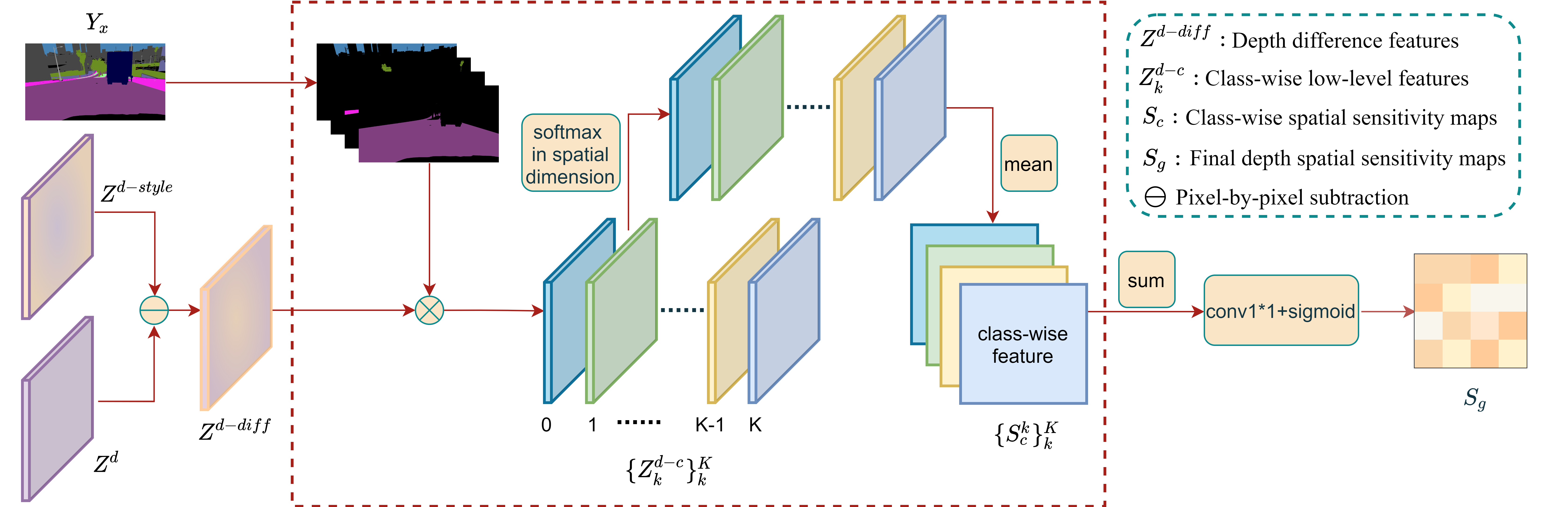}
	\caption{The illustration of the proposed class-wise soft spatial sensitivity suppression. $Z^{d-style}$, $Y_{x}$ and $Z^{d}$ are consistent with those mentioned above.  }
	\label{css}
\end{figure*}
\subsection{Class-wise soft spatial sensitivity suppression}
To augment the non-sensitive features of depth, the class-wise soft spatial sensitivity suppression is designed as presented in Fig. \ref{css}. Given an original depth feature $Z^{d}$, the stylized depth feature $Z^{d-style}$ is acquired through the RGB-D inter-modal stylization flow. The stylized depth feature obtains some characteristics of RGB while maintaining the original traits of the depth feature. Afterward, the feature difference $Z^{d-diff}$ caused by the stylization flow is computed which is defined as: 
\begin{equation}
	\label{deqn_ex1a}
	Z^{d-diff} = |Z^{d} -   Z^{d-style}|
\end{equation}

Feature sensitivity is typically calculated from a channel perspective \cite{xu2022dirl}, but for a single-channel depth map, the information contained in the channel is considered limited compared to the spatial information. Moreover, sensitivity is generally calculated globally \cite{choi2021robustnet}, \cite{wu2022siamdoge}, which results in certain class-specific sensitive features being overlooked. Furthermore, high-sensitivity hard suppression, due to its binarization operation, struggles to capture continuous changes and detailed information in sensitive regions. Additionally, it lacks flexibility in complex scenarios. Therefore, a soft spatial sensitivity suppression is employed at the class level in this work to strengthen non-sensitive features for better generalization.  

Firstly, the difference feature $Z^{d-diff}\in \mathbb{R}^{B, C, H, W} $ is leveraged to generate the k-class features ${Z}_{k}^{d-c}\in \mathbb{R}^{B, C, H, W}$ based on the labels $Y_{x}$, where $k\in K $. Second, the spatial elements of class-wise features across each channel are normalized by Softmax operation, ensuring that the sum of pixel probabilities for that class in the channel is 1. Then, the class-wise spatial sensitivity map $S_{c}^{k}\in \mathbb{R}^{B, 1, H, W}$ is derived by performing a mean operation across the channels. After summing all class-wise sensitivity maps pixel by pixel, a 1$\times$1 convolution and sigmoid function are applied for normalization, leading to an efficient final global sensitivity map $S_{g}\in \mathbb{R}^{B, 1, H, W}$. The more sensitive a feature is, the higher weight it is given. Then the non-sensitive map $N_{g}\in \mathbb{R}^{B, 1, H, W} $ is derived by subtracting $S_{g}$ from an all-ones tensor $\mathbf{1}\in \mathbb{R}^{B, 1, H, W}$. Formally, the process is:
\begin{equation}
	\label{deqn_ex1a}
	S_{c}^{k} =  Mean ( Softmax (Z_{k}^{d-c}) )
\end{equation}
\begin{equation}
	\label{deqn_ex1a}
	S_{g} = Sigmoid(Conv_{1\times1}(\sum_{k=0}^{K} S_{c}^{k}))
\end{equation}
\begin{equation}
	\label{deqn_ex1a}
	N_{g} = \mathbf{1} - S_{g}
\end{equation}

Unlike classic methods \cite{choi2021robustnet}, \cite{wu2022siamdoge} that suppress sensitive features via loss functions, the proposed DSSS approach directly strengthens insensitive features using a non-sensitivity map. The refined depth feature $Z^{d-fine}$ is computed as:

\begin{equation}
	\label{deqn_ex1a}
	Z^{d-fine} = Z^{d} \odot N_{g} + Z^{d}
\end{equation}

Here, The element-wise multiplication $\odot$ selectively amplifies the insensitive features that contain more domain-invariant information, while the element-wise addition preserves the overall structure and inherent properties of the depth map.

Subsequently, the same operation is applied to fuse the RGB low-level feature $Z^{rgb}$ and the refined depth feature $Z^{d-fine}$, yielding more robust features for improved generalization. The fusion process is mathematically described as follows:
\begin{equation}
	\label{deqn_ex1a}
	Z^{rgb-d} = Z^{d-fine} \odot Z^{rgb} + Z^{rgb}
\end{equation}
\subsection{RGB-D soft alignment loss}
The fundamental objective of the DG algorithm is to capture domain-invariant features to the greatest extent. RGB and stylized depth images offer learnable domain-invariant features, or more accurately, modality-invariant ones. Considering that RGB and depth only share some modality-specific features, introducing the soft alignment loss $ \mathcal{L}_{SA}$ between the stylized depth features and the original RGB features becomes necessary. This allows both modalities to maintain some similarity while preserving their unique characteristics. Formally, this slack relationship is maintained through the following loss function:
\begin{equation}                                                 
	\label{deqn_ex1a}
	\mathcal{L}_{SA} = \beta\frac{1}{N} \sum_{i=1}^{N} \left( {Z^{rgb}_{i}- Z^{d-style}_{i}} \right)^2
\end{equation}
where $N$ is the total number of pixels, $i$ represents the pixel at a specific position and $\beta$ is a scaling factor that controls the influence of the RGB features on the stylized depth maps.

\section{Experiments}
In this part, extensive experiments are performed to demonstrate the effectiveness of our proposed approach. Five datasets, including GTA5 \cite{richter2016playing}, SYNTHIA \cite{ros2016synthia}, Cityscapes \cite{cordts2016cityscapes}, SELMA \cite{testolina2023selma}, and InfraParis \cite{franchi2024infraparis}, are utilized to examine the generalization ability of our framework. Comparisons are made with the latest state-of-the-art approaches, and ablation studies for each component are included. Furthermore, implementation details and qualitative visualizations are also presented. 
\subsection{Datasets}
Our approach is validated on five datasets: the synthetic datasets GTA5\cite{richter2016playing}, SYNTHIA\cite{ros2016synthia}, and SELMA\cite{testolina2023selma}, and the real-world datasets Cityscapes\cite{cordts2016cityscapes} and InfraParis \cite{franchi2024infraparis}. The labels for these datasets follow the annotation format used by Cityscapes (19 classes).
\begin{itemize}
	\item \textbf{GTA5 dataset} comprises 24,966 images at a resolution of 1914×1052, divided into 12,403 training images, 6,382 validation images, and 6,181 testing images. Following CorDA \cite{wang2021domain}, the same Monodepth2 model \cite{godard2019digging} is used to render the depth maps for the GTA5 dataset.
	\item \textbf{SYNTHIA dataset} holds 9,400 images with a solution of 1280×760, of which 6,580 designated for training and 2,820 for validation. Notably, both the depth maps and labeled images are generated together using a rendering engine.
	\item \textbf{SELMA dataset} comprises 30,909 photo-realistic on-road images with a resolution of 1280×640, gathered from 216 different conditions. A total of 16,675 images are used for training, 7,035 for validation, and 7,199 for testing. In addition, each RGB image is paired with a corresponding semantic annotation and depth map.
	\item \textbf{Cityscapes dataset} collected in European cities, is one of the most popular driving datasets for urban scenes. A total of 2,975 training images and 500 validation images, with a resolution of 2048×1024, are provided and labeled for 19 classes. Stereo depth estimations, sourced from \cite{godard2019digging}, are created through Semi-Global Matching \cite{hirschmuller2005accurate} and stereoscopic inpainting \cite{wang2008stereoscopic}. As depth maps are exclusively available in the training set, our validation experiments will be performed on the Cityscapes dataset using this set, with the same procedure applied to other comparison approaches.
	\item \textbf{InfraParis dataset} captured in various areas around Paris, consists of 7,301 samples, divided into 6,567 training images, 189 validation images, and 571 test images. It supports various tasks across three distinct modalities: RGB, depth, and infrared. In our experiments, the validation and test images are merged as the validation dataset. Additionally, due to the fact that the front of the vehicle occupies more than a third of the image in the collected photos, the front of the car is cropped to obtain more effective results.
\end{itemize}

\subsection{Implementation details}
Our experiments are conducted by adopting DeepLabV3+ \cite{chen2018encoder} with ResNet-50 \cite{he2016deep}, ShuffleNet-V2 \cite{ma2018shufflenet}, and MobileNet-V2 \cite{sandler2018mobilenetv2} backbones as segmentation networks. All of the backbones have been initialized using the ImageNet pre-trained weights. Prior to integrating RGB-D fusion, MobileNet-V2 is consistently utilized to extract depth low-level features with little additional computational cost. The input images are cropped to 768×768 resolution, and a batch size of 4 is used. The SGD optimizer starts with a learning rate of 0.01 and a momentum of 0.9, while the learning rate is reduced by a factor of 0.9 according to the polynomial scheduler. All experiments are conducted on a single NVIDIA GeForce RTX3090Ti GPU, with the maximum number of iterations set to 40k. The weight of RGB-D soft alignment loss is set to 0.1 and the K equals to 19. The mean intersection-over-union (mIoU) \cite{everingham2015pascal} is consistently adopted as the performance evaluation metric in all experiments. 

\begin{table*}[t]
	\caption{Performance comparison with other methods using heavyweight backbones. \textbf{Bold} format and \uline{Underline} represent the best and the second-best performance, respectively. Mean indicates the average performance for each source domain, and Avg reflects the total average performance across the two source domains. * implies that the external dataset (e.g., ImageNet) in WildNet is substituted with the source dataset for fairness in evaluation.}
	\label{tab: compare to RGB DG performance }
	\centering{}\resizebox{1.0\textwidth}{!}{%
		\begin{tabular}{ccccccc|ccccc|ccccc}
			\toprule 
			\multirow{2}{*}{Model} &\multirow{2}{*}{Methods} &\multirow{2}{*}{RGB}&  \multirow{2}{*}{Depth} & \multirow{2}{*}{FLOPs (G)} &\multirow{2}{*}{Params (M)} &  \multirow{2}{*}{Avg}& \multicolumn{5}{c}{Trained on GTA5 (G)} & \multicolumn{5}{c}{Trained on Cityscapes (C)}  \\
			&&&& && & $\rightarrow$C & $\rightarrow$SYN & $\rightarrow$SEL & $\rightarrow$I & $\rightarrow$Mean & $\rightarrow$G & $\rightarrow$SYN & $\rightarrow$SEL & $\rightarrow$I & $\rightarrow$Mean  \tabularnewline 
			\hline 
			\noalign{\vskip0.1cm}
			&DeepLabV3+ \cite{chen2018encoder} &\checkmark  & &139.05 &45.08   &36.04   &30.71  &26.75  &33.25  &32.23   &30.73  &42.84 &23.23  &39.26  &60.07   &41.35  \tabularnewline
			&IBN \cite{pan2018two} &\checkmark  &&139.05 &45.08   &38.08    &34.39  &27.81 &34.40  &35.20   &32.95  &$\underline{44.62}$  &$\underline{26.02}$  &39.94 
			&\textbf{62.22}  &$\underline{43.20}$   \tabularnewline
			&ISW \cite{choi2021robustnet} &\checkmark  & &139.05 &45.08   &39.07    &38.05  &27.25  &36.91  &39.43  &35.41   &44.17  &25.84  &40.30   &$\underline{60.60}$  & 42.73   \tabularnewline
			ResNet-50&WildNet* \cite{lee2022wildnet} &\checkmark  & &139.01 &45.21    &$\underline{39.54}$  &$\underline{39.90}$  &27.36  &$\underline{37.06}$  &\textbf{42.24}  &$\underline{36.64}$  	&\textbf{45.44} &25.06  &39.53  	&59.71  	&42.43  \tabularnewline
			
			&DGFD \cite{liu2024segment} &\checkmark  &\checkmark   &384.27   &94.15  &28.45  &27.09  &22.92  &27.23  &27.15  &26.10  &31.09  &21.10  &27.33 &43.70 &30.80  \tabularnewline
			
			&Ours & \checkmark &\checkmark &139.54 &45.14  &\textbf{41.47}   &\textbf{42.07}  &\textbf{31.80} 	&\textbf{38.28} &$\underline{42.05}$ &\textbf{38.55}  &44.04  	&\textbf{31.82}  	&$\underline{42.09}$  &59.61  & \textbf{44.39}  \tabularnewline
			\hline 
			\noalign{\vskip0.1cm}
			MiT-B2&CMX \cite{zhang2023cmx} &\checkmark  &\checkmark   &114.28 &66.57 & 36.24   &28.19  &$\underline{28.74}$  &33.78  &28.72 &29.86  &43.54  &25.69  &\textbf{43.44}  &57.81  &42.62     \tabularnewline   
			\hline
			\noalign{\vskip0.1cm}
			
	\end{tabular}}
	
\end{table*}

\subsection{Comparative Results}
\subsubsection{Comparison With Other DG Methods Only Using RGB}

\par Some state-of-the-art DG approaches are compared with our proposed approach as shown in Table  \ref{tab: compare to RGB DG performance }  and Table \ref{tab:DG_COMPARE_to_Other_LIGHTWEIGHT}, which include DeepLabV3+\cite{chen2018encoder}, IBN-Net\cite{pan2018two}, ISW\cite{choi2021robustnet}, WildNet\cite{lee2022wildnet}. To ensure fair comparisons, most of methods, opening their source code, are re-implemented and evaluated on the datasets. 

Based on the heavyweight backbones, our approach outperforms the second-best method, WildNet, with a 1.93\% increase in average mIoU across two tasks. Among eight validation experiments, our method achieves first place in four cases and second place in two. Our method, using G as the training set, outperforms the second-best with gains of 2.17\%, 3.06\%, and 1.22\% on Cityscapes, SYNTHIA, and SELMA datasets. With C, it achieves 5.80\% improvements on SYNTHIA dataset.
\begin{figure*}[t]
	\centering
	\includegraphics[width=1\textwidth]{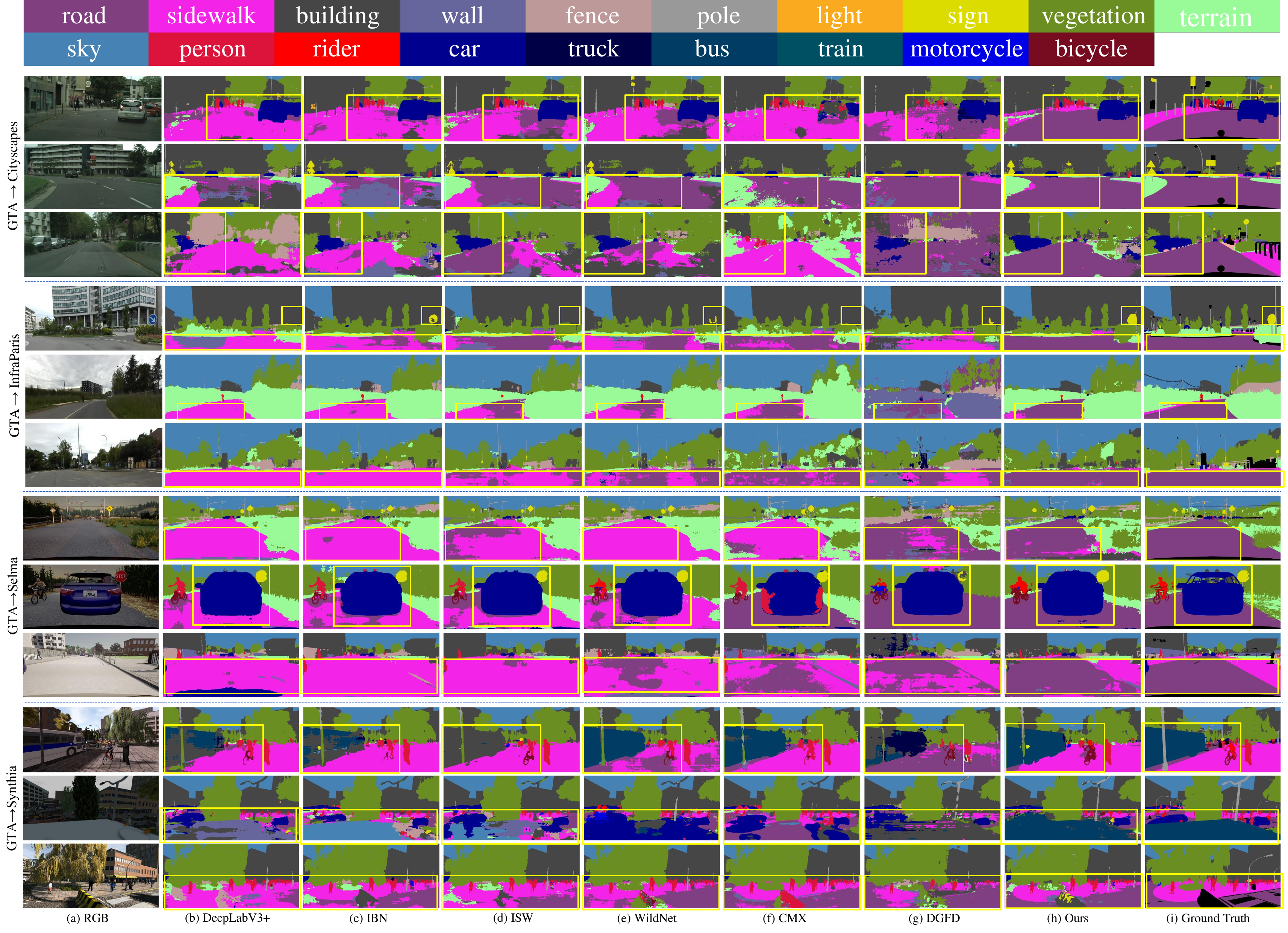}
	\caption{Visualization comparison between our DSSS approach and state-of-the-art methods  for the GTA5-to-other-datasets generalization task using heavyweight backbones.}
	\label{compare2Others_resnet}
\end{figure*}
Based on the ShuffleNet-V2 and MobileNet-V2 backbones, 1.68\% and 2.18\% improvement in the average mIoU are achieved when compared to the second-best approach. Both lightweight backbones achieve the best performance across all validation experiments. For the task of GTA5 generalizing to Cityscapes, our approach achieves 35.32\% and 33.94\% average mIoU with ShuffleNet-V2 and MobileNet-V2 backbones, respectively, outperforming other state-of-the-art methods and achieving gains of 8.11\% and 7.77\% over the baseline.
\begin{figure*}[t]
	\centering
	\includegraphics[width=1\textwidth]{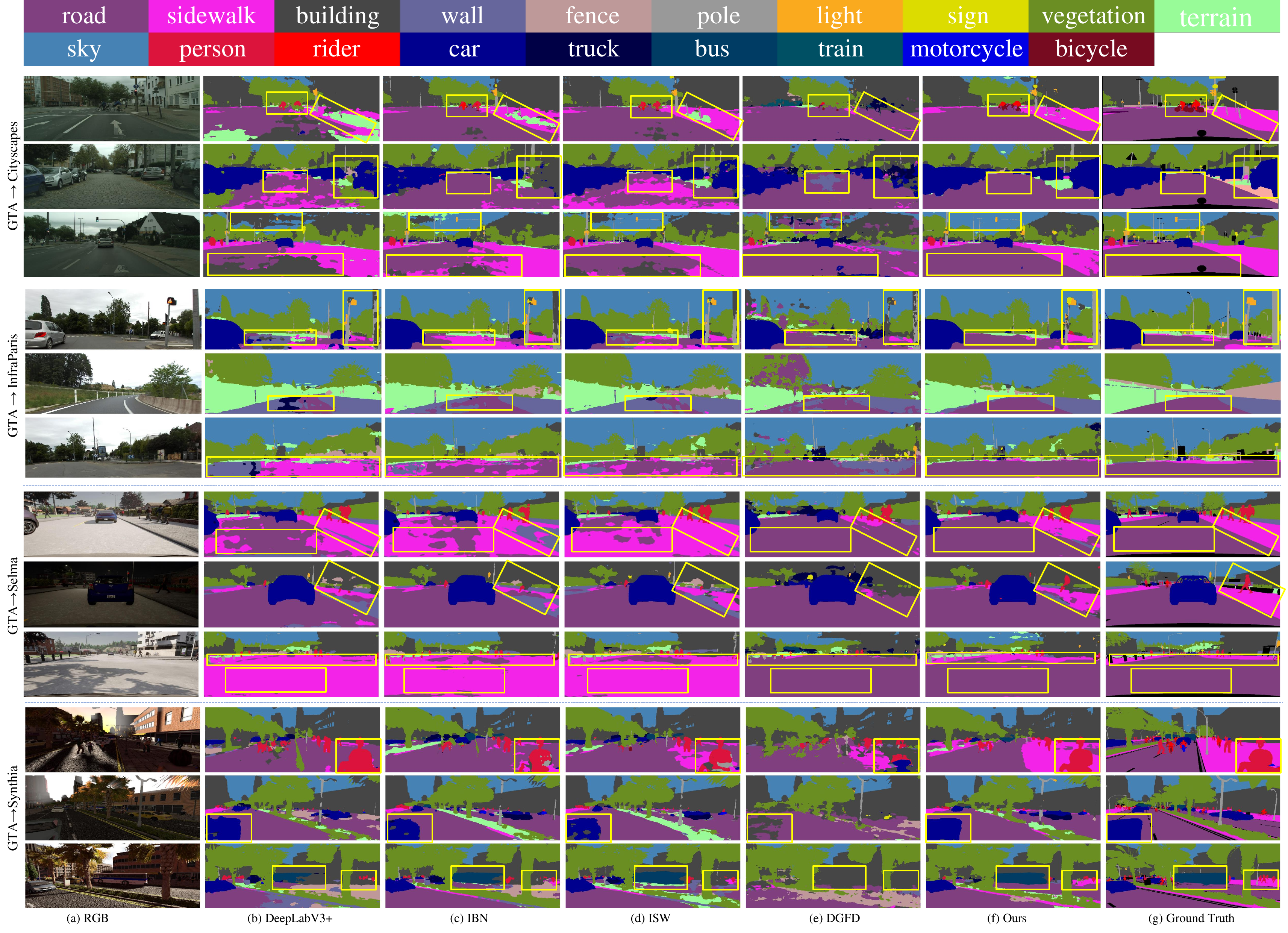}
	\caption{Visualization comparison between our DSSS approach and state-of-the-art methods for the GTA5-to-other-datasets generalization task using ShuffleNet-V2 backbone.}
	\label{compare2Others_shuffleNet}
\end{figure*}
\begin{table*}[t]
	\caption{Performance comparison with other methods using lightweight backbones including ShuffleNet-V2 and MobileNet-V2, on the G $\rightarrow$ \{C, SYN, SEL, I\} generalized task. \textbf{Bold} format and \uline{Underline} represent the best and the second-best performance, respectively. Mean indicates the average performance for the source domain.}
	\label{tab:DG_COMPARE_to_Other_LIGHTWEIGHT} 
	\centering{}\resizebox{1.0\textwidth}{!}{%
		\begin{tabular}{ccccc|cccc}
			\toprule 
			\multirow{2}{*}{Model} & \multirow{2}{*}{Method} & \multirow{2}{*}{FLOPs (G)} &\multirow{2}{*}{Params (M)} &  \multirow{2}{*}{Avg} & \multicolumn{4}{c}{Trained on G} \\
			&&&&& $\rightarrow$C&  $\rightarrow$SYN &  $\rightarrow$SEL  &$\rightarrow$I \tabularnewline
			\cline{1-9}
			\noalign{\vskip0.1cm}
			\multirow{5}{*}{ShuffleNet-V2} 
			& DeepLabV3+ \cite{chen2018encoder} &29.41 &12.64 &27.20 &27.21  &23.46  &28.55  &29.60  \tabularnewline
			& IBN\cite{pan2018two} &29.41 &12.64 &29.28 &27.87  &$\underline{25.30}$  &30.38  &33.58  \tabularnewline
			& ISW \cite{choi2021robustnet}  &29.41 &12.64 &$\underline{30.70}$ &$\underline{33.25}$  &24.47  &$\underline{31.33}$  &$\underline{33.74}$ \tabularnewline
			&DGFD\cite{liu2024segment}  &64.59 &23.03 &22.97 &23.39 &18.61 &22.91 &26.97   \tabularnewline
			& Ours &29.74 &12.70 &\textbf{32.38} &\textbf{35.32}  &\textbf{28.04}  &\textbf{31.74}  &\textbf{34.42}  \tabularnewline
			\cline{1-9}
			\noalign{\vskip0.1cm}
			\multirow{5}{*}{MobileNet-V2}
			&DeepLabV3+ \cite{chen2018encoder} &36.05 &14.84  &27.15 &26.17 & 23.62 &29.38  &29.42 \tabularnewline
			& IBN\cite{pan2018two} &36.05 &14.84 & 28.12 &29.30  &$\underline{24.52}$  & 28.02 &30.66 \tabularnewline
			& ISW \cite{choi2021robustnet} &36.05 &14.84 &$\underline{30.04}$  &$\underline{32.72}$  &24.21  &$\underline{29.80}$  &$\underline{33.42}$  \tabularnewline
			&DGFD \cite{liu2024segment}  &109.57 &32.15 &21.24 &21.00 &18.59 &21.90 &23.48   \tabularnewline
			&Ours  &36.37 &14.90 &\textbf{32.22} &\textbf{33.94}  &\textbf{28.50}  &\textbf{32.69} &\textbf{33.74} \tabularnewline
			\hline
			\noalign{\vskip0.1cm}
	\end{tabular}}
\end{table*}

In summary, our model achieves the best performance regardless of whether it uses a heavyweight backbone network or a lightweight backbone network, which once again proves that depth maps are indeed helpful for cross-domain learning. Moreover, Fig. \ref{compare2Others_resnet} and Fig. \ref{compare2Others_shuffleNet} respectively show the visualization results of the model trained on the synthetic GTA5 dataset using a heavyweight backbone network and a lightweight backbone network represented by ShuffleNet-V2 to generalize to other datasets. The area with the most obvious effect has been marked with a yellow rectangle. From two visualization results, it is evident that our model achieves remarkable performance across datasets. Specifically, while most RGB-based methods misclassify roads as sidewalks during segmentation, our approach successfully differentiates between these classes. 	

\subsubsection{Comparison With Other Methods Using RGB-D}
To further validate the effectiveness of our approach, two RGBD methods, CMX\cite{zhang2023cmx} and DGFD\cite{liu2024segment}, are implemented for comparison under two generalization settings. CMX focuses on utilizing transformer architectures to explore the possibilities of RGB-D fusion. To maintain fairness, CMX experiments are limited to the MiT-B2 backbone, while DGFD and our experiments are conducted on three distinct backbones.

As presented in Tables \ref{tab: compare to RGB DG performance } and \ref{tab:DG_COMPARE_to_Other_LIGHTWEIGHT}, our approach achieves the highest performance across both heavyweight and lightweight backbones across two settings. Specifically, in the heavyweight network, our method surpasses CMX and DGFD by 5.23\% and 13.02\%, respectively. Notably, using GTA5 as the training set yields the most significant performance gains, with increases of 8.69\% and 12.45\%, respectively. The drop performance of CMX highlights the limited generalization ability of supervised semantic segmentation methods. This issue is particularly prominent when transferring knowledge learned from synthetic datasets to other domains, especially real-world datasets, where performance suffers considerably. However, when trained on Cityscapes, CMX achieves impressive generalization results, largely owing to the robust capacity of transformer to generalize and extract meaningful insights from real-world datasets, thereby enhancing performance on other datasets. In contrast, the performance degradation of DGFD can be attributed to its binary-classification-focused design, which inherently limits its ability to handle the complexity of multi-class semantic segmentation tasks. In the lightweight network, our method outperforms DGFD by 9.41\% and 10.98\% when using ShuffleNet-V2 and MobileNet-V2 backbones. The visualization result are shown in Fig. \ref{compare2Others_resnet} and Fig. \ref{compare2Others_shuffleNet}.

\subsubsection{Model Complexity Analysis}
The computational cost of the proposed approach is analyzed during the inference stage, with the input size set to 3 × 1024 × 512. As shown in Table \ref{tab: compare to RGB DG performance } and Table \ref{tab:DG_COMPARE_to_Other_LIGHTWEIGHT}, which detail the computational costs and parameter counts, our method delivers performance improvements while keeping computational costs and parameter counts nearly identical to other RGB-based methods on both heavyweight and lightweight backbones. In contrast to RGB-D methods, our approach not only drastically reduces both computational cost and parameter count but also enhances generalization performance significantly. Specifically, DGFD exhibits the largest FLOPs and parameter counts among both heavyweight and lightweight backbones, with its heavyweight configuration reaching 384.27G FLOPs and 94.15M parameters.

\begin{table*}[t]
	\caption{Ablation studies on the proposed components using ResNet-50 on the task of G generalizing to other datasets.}
	\label{tab:ablation}
	\centering{}\resizebox{1.0\textwidth}{!}{%
		\begin{tabular}{cccc|cccccc|ccccc}
			\toprule 
			Backbone &Group &RGB&  Depth&  GCSS &CHSS &CSSS  &  RSM &IMSF &SAL & $\rightarrow$C & $\rightarrow$SYN & $\rightarrow$SEL & $\rightarrow$I & $\rightarrow$Mean  \tabularnewline 
			\hline 
			\noalign{\vskip0.1cm}
			\multirow{6}{*}{ResNet-50} 
			&A  & \checkmark &   & & & & &&   &34.39  &27.81 &34.40 &35.20 &32.95 \tabularnewline
			&B  & \checkmark & \checkmark  && & & & &   & 37.99 &30.06 &37.66 &37.39 &35.78 \tabularnewline
			&C  & \checkmark &\checkmark   &\checkmark & &&\checkmark & &   &38.62  &31.17 &35.58 &40.13 &36.38 \tabularnewline
			&D  & \checkmark &\checkmark   && &\checkmark &\checkmark & &   & 38.86 &31.23 &37.85 &41.03 &37.24 \tabularnewline
			&E  & \checkmark &\checkmark   & &\checkmark& &\checkmark & &   &38.06  &29.72 &34.01 &35.72 &34.38 \tabularnewline
			&F  & \checkmark &\checkmark   & &&\checkmark & &\checkmark &   &41.23  &\textbf{32.61} &37.94 &41.38 &38.29 \tabularnewline
			&G  & \checkmark &\checkmark   & &&\checkmark & &\checkmark & \checkmark  & \textbf{42.07} &31.80 & \textbf{38.28}&\textbf{42.05} &\textbf{38.55} \tabularnewline
			
			\hline
			\noalign{\vskip0.1cm}
	\end{tabular}}
\end{table*}
\begin{figure*}[h]
	\centering
	\includegraphics[width=1\textwidth]{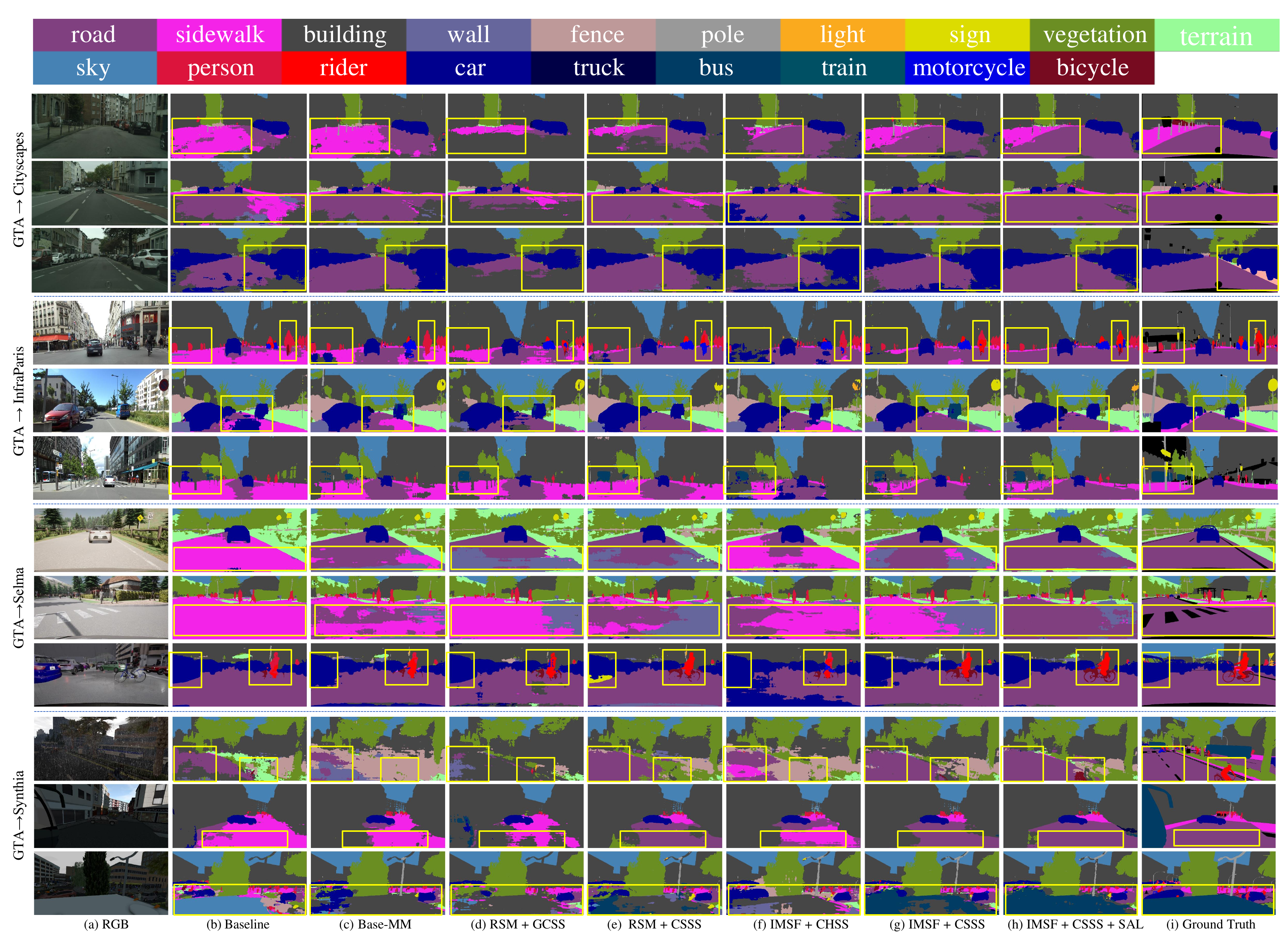}
	\caption{\centering{Visualization results of ablation studies for the GTA5-to-other-datasets generalization task of our proposed method using ResNet-50 backbone.}}
	\label{ablated_visualization}
\end{figure*}
\subsection{Ablation Studies}
\par As illustrated in Table \ref{tab:ablation}, the effectiveness of the three proposed components, CSSS, IMSF, and SAL, is investigated. Experiments are carried out using the ResNet-50 backbone. The GTA5 dataset is selected as the source domain, while the other four datasets are used as unseen target domains. Additionally, GCSS stands for calculating the soft sensitivity in channel dimension globally, CHSS represents the threshold-based calculation of hard sensitivity in the spatial dimension, and RSM means obtaining the stylized depth image using methods like random colorization and blur. 

Group A corresponds to the baseline model, while Group B represents the base-MM model. The base-MM model fuses low-level RGB and depth features by first multiplying them and then adding the result to the original RGB features. As shown in Table \ref{tab:ablation}, the performance with the RGB-D fusion module reaches an mIoU of 35.78\%, representing a gain of no less than 2.83\% over the baseline model. The findings of this experiment confirm the theory that extra geometric information is contributed by depth images to RGB, exhibiting complementary features. Based on Group B, Group F integrates the CSSS and IMSF to bolster depth-insensitive features, achieving a 5.34\% increase in average mIoU compared to the baseline. Group G, which uses all the proposed components, provides the greatest improvement, achieving 38.55\% in average mIoU across the four datasets.

Group C integrates the GCSS and RSM, leading to a 0.6\% improvement in average mIoU over base-MM. Similarly, Group D adopts the CSSS and RSM, achieving a 1.46\% increase in mean mIoU compared to base-MM. Both groups effectively demonstrate that the sensitivity suppression strategy enhances the learning of domain-invariant features in depth maps, thereby strengthening the overall generalization ability of the model. However, when comparing Group C and Group D, it is evident that the class-wise soft spatial sensitivity suppression yields better results. This finding supports spatial sensitivity plays a more significant role than channel sensitivity for depth features. To further validate the effectiveness of CSSS, Group E conducts experiment revealing the limitations of threshold-based hard spatial sensitivity methods in capturing continuous depth variations and fine details on depth maps. Specifically, Group E sets the sensitivity threshold to 20, and when the depth class-wise difference exceeds this threshold, the corresponding positions in the sensitivity matrix are set to 1. Otherwise, they are set to 0. The experimental results shows that the model achieved an mIoU value of 34.38\%, which is a 1.40\% decrease compared to the base-MM model and a 2.86\% decrease compared to Group D. This phenomenon suggests that the high-sensitivity hard suppression method overly simplifies the continuity and fine details of the depth map, potentially leading to information loss and significantly affecting the degradation performance. These results further highlight the advantages of the high-sensitivity soft suppression method (CSSS) in preserving depth map details and continuity.

When comparing Group A, D, and A, F, the model achieves a performance improvement of 4.29\% and 5.34\% in average mIoU, and a 1.05\% increase is observed between Group D and F. These gains demonstrate the effectiveness of the IMSF, showing that shared representations exist between RGB and stylized depth, and that modality-invariant features can be transferred by IMSF. Furthermore, a 5.6\% enhancement in average mIoU is observed on the four validation datasets when leveraging all three proposed components compared to the baseline model, ranging from 32.95\% to 38.55\%. The clear increase in performance illustrates the effectiveness of the proposed components.

Similarly, Fig. \ref{ablated_visualization} also provides the visualization results of our ablation experiments under the GTA5-to-other-datasets experimental setup, with ResNet-50 as the backbone. Consistent with our experimental findings, the results indicate that our method achieves superior segmentation, particularly excelling in object boundary segmentation.

\subsection{Qualitative Analysis}
\subsubsection{Class-wise Performance Analysis}
\begin{figure*}[t]
	\centering
	\includegraphics[width=1\textwidth]{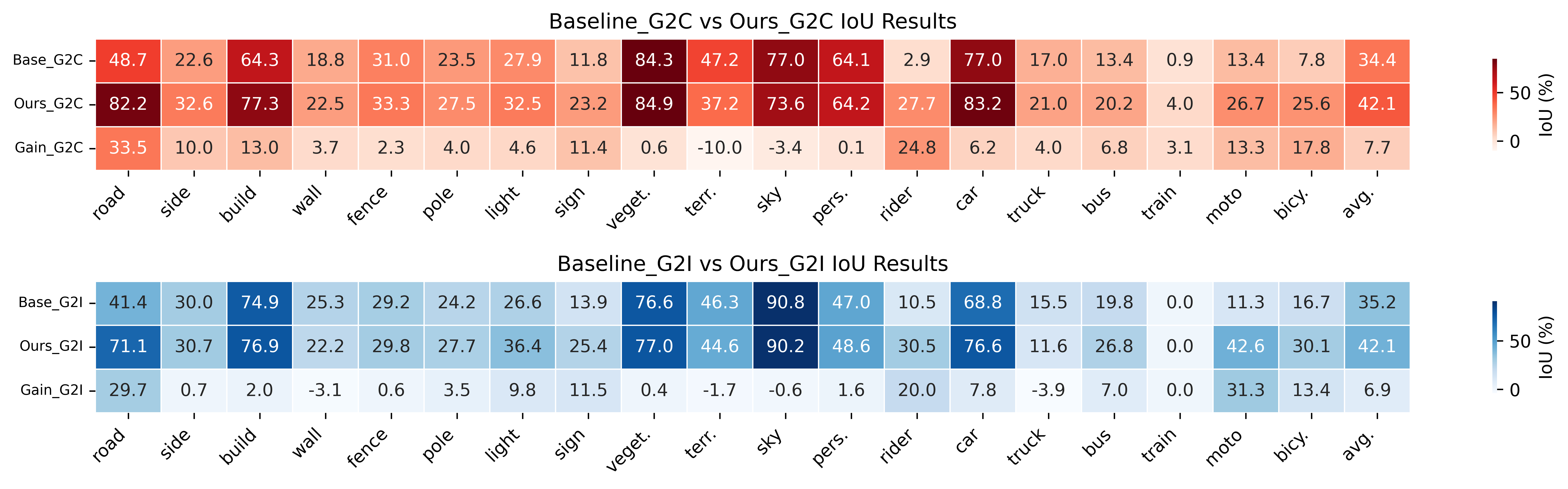}
	\caption{
		Heatmap visualization of the IOU results and gains of the baseline and our model for 19 classes on two real-world validation datasets Cityscapes and InfraParis.}
	\label{heatmap}
\end{figure*}

To further validate the effectiveness of the proposed method, both the baseline and the proposed model are trained using the ResNet-50 backbone on the GTA5 dataset, and evaluated on two real-world datasets (C and I). Additionally, heatmaps of the IoU values and gains for the 19 categories are generated, as shown in Fig. \ref{heatmap}. The higher the value, the deeper the color. The results demonstrate that the proposed method achieves performance improvements across nearly all categories in both datasets, particularly in the \textquotedblleft rider\textquotedblright, \textquotedblleft traffic sign\textquotedblright, \textquotedblleft traffic light\textquotedblright, \textquotedblleft motorcycle\textquotedblright, and \textquotedblleft bicycle\textquotedblright\ categories, where substantial gains are observed. Notably, the IoU for the \textquotedblleft rider\textquotedblright\ category improves by 24.8\% and 20.0\% in Cityscapes and InfraParis, respectively. These results suggest that the proposed model effectively enhances performance in handling rare classes.

\begin{table}[h]
	\caption{Performance comparison with other methods using three backbones in the low-light environment. \textbf{Bold} format and \uline{Underline} represent the best and the second-best performance, respectively. Avg reflects the total average performance across the three backbones.}
	\label{tab: night DG performance }
	\centering{}\resizebox{1\textwidth}{!}{%
		\begin{tabular}{cccc|ccc}
			\toprule 
			Methods &RGB &Depth &Avg  &ResNet-50  & ShuffleNet-V2 & MobileNet-V2 \tabularnewline 
			\hline 
			\cline{1-7}
			\noalign{\vskip0.1cm}
			DeepLabV3+ \cite{chen2018encoder} &\checkmark  &  &21.09 &25.99   &18.52  &18.74 \tabularnewline
			IBN \cite{pan2018two} &\checkmark  &  &25.23 &28.37     &$\underline{24.41}$ &22.90 \tabularnewline
			ISW \cite{choi2021robustnet} &\checkmark &  &$\underline{26.39}$ &$\underline{31.22}$ &24.32  &$\underline{23.63}$   \tabularnewline
			DGFD \cite{liu2024segment} &\checkmark  &\checkmark   &14.60 &15.22    &15.11  &13.45   \tabularnewline
			Ours & \checkmark &\checkmark &\textbf{28.42} 	&\textbf{32.96}  &\textbf{25.71}  	&\textbf{26.58}  \tabularnewline	
			\hline
			\noalign{\vskip0.1cm}
			
	\end{tabular}}
\end{table}

\begin{figure*}[h]
	\centering
	\includegraphics[width=0.8\textwidth]{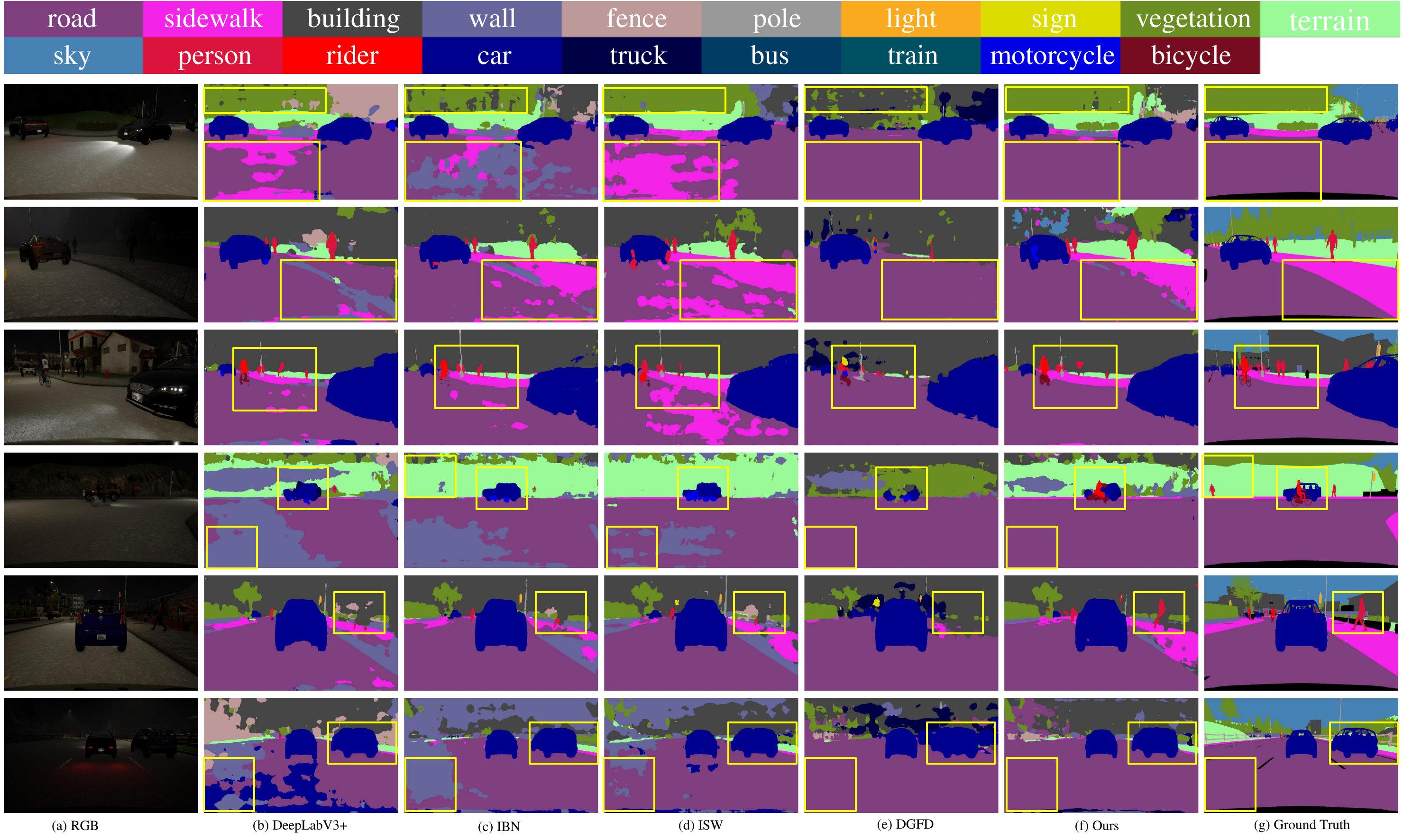}
	\caption{Visualization comparison of our DSSS approach and state-of-the-art methods on the GTA5-to-NightSelma generalization task, utilizing the ShuffleNet backbone in a low-light environment.}
	\label{night}
\end{figure*}

\subsubsection{Comparative Analysis in low-light Scenarios}
To verify the performance of our proposed model in low-light environments, a low-light validation dataset, NightSelma, is constructed by extracting photos generated from night environment in various weather conditions from the SELMA validation set, comprising a total of 1,780 images. Table \ref{tab: night DG performance } presents a comparison of several state-of-the-art methods with the proposed approach. Among the three different backbone networks, our method achieves the best performance, with an average performance of 28.42\%, which is 2.03\% higher than the second best method. Among them, the best performance is the MobileNet-V2 backbone network, which is 2.95\% higher than the second best method. As shown in Fig. \ref{night}, several evaluated images from NightSelma dataset are selected for visualization. All compared methods including ours are trained on the GTA5 dataset using the ShuffleNet-V2 backbone. The results indicate that, in low-light environments, methods relying solely on RGB images sometimes struggle to clear the boundary between the sidewalk and the road, and fail to fully recognize the shape of small objects. For DGFD, which uses RGB and depth as input, it performs better than the methods relying solely on RGB when it comes to segmenting low-light roads. However, since it is originally designed only for segmenting roads and backgrounds, it is obviously weak in other categories. Our method effectively alleviates these two problems.

\section{Conclusion}
In this paper, based on the novel perspective that depth maps are not always reliable in providing domain-invariant features, we explore the potential of depth-sensitive soft suppression in improving the performance of DGSS tasks. To the best of our knowledge, this is the first work focusing on RGB-D data to solve multi-class DGSS problem. Specifically, we propose the RGB-D inter-modal stylization flow and RGB-D soft alignment loss to cleverly utilize RGB information to stylize depth maps for sensitivity detection, and make the stylized depth maps retain part of the RGB information and preserve its essential characteristics. In addition, we propose the class-wise soft spatial sensitivity suppression to identify and emphasize non-sensitive regions in depth maps that contain domain-invariant features for each class. Extensive experiments demonstrate the superior generalization ability of our model, even under low-light conditions.






\bibliographystyle{elsarticle-num-names} 
\bibliography{ref}


%
\end{document}